\documentclass[conference]{IEEEtran}
\IEEEoverridecommandlockouts
\usepackage{cite}
\usepackage{amsmath,amssymb,amsfonts}
\usepackage{algorithmic}
\usepackage{graphicx}
\usepackage[caption=false,font=normalsize,labelfont=sf,textfont=sf]{subfig}
\usepackage{textcomp}
\usepackage[table]{xcolor}
\usepackage{colortbl}
\usepackage{tcolorbox}
\usepackage{tikz}
\usepackage{fancyhdr}
\usepackage{eso-pic}

\def\BibTeX{{\rm B\kern-.05em{\sc i\kern-.025em b}\kern-.08em
    T\kern-.1667em\lower.7ex\hbox{E}\kern-.125emX}}
\definecolor{bestgreen}{RGB}{198,239,206}
\definecolor{goodyellow}{RGB}{255,235,156}
\definecolor{badred}{RGB}{255,199,206}

\begin{document}

\newcommand{\IEEEcopyrightbox}{
\AddToShipoutPictureFG*{
\put(57,34){
\begin{minipage}{500pt}
\begin{tcolorbox}[
    colback=white,
    colframe=black,
    boxrule=0.5pt,
    arc=0pt,
    left=4pt,
    right=4pt,
    top=3pt,
    bottom=3pt
]
\centering
\small
© 2026 IEEE. Personal use of this material is permitted. Permission from IEEE must be obtained for all other uses, including \\
reprinting/republishing this material for advertising or promotional purposes, collecting new collective works for resale or \\
redistribution to servers or lists, or reuse of any copyrighted component of this work in other works.
\end{tcolorbox}
\end{minipage}
}
}
}

\title{Continuous-Space Roadmap Generation for Mobile Robot Fleets with Distance Constraints and Geometry-Aware Discretization}

\author{
\IEEEauthorblockN{Marvin Rüdt$^1$$^,$$^2$}
\IEEEauthorblockA{
marvin.ruedt@kit.edu}
\and
\IEEEauthorblockN{Constantin Enke$^1$}
\IEEEauthorblockA{
constantin.enke@kit.edu}
\and
\IEEEauthorblockN{Kai Furmans$^1$}
\IEEEauthorblockA{
kai.furmans@kit.edu}
\thanks{$^1$Institute for Material Handling and
Logistics, Karlsruhe Institute of Technology,
Karlsruhe, Germany.}
\thanks{$^{2}$Corresponding author.}
}

\maketitle

\IEEEcopyrightbox

\begin{abstract}
Efficient routing of mobile robot fleets requires roadmaps with high redundancy, short path lengths, and sufficient node and edge clearance for conflict-free operation. Existing grid-based methods sacrifice geometric fidelity and impose Manhattan-distance path length constraints, whereas existing continuous-space methods neglect minimum distance constraints and transport demand. This paper proposes a continuous-space roadmap generation method that addresses this gap by placing nodes at convex corner points of the free space and at station interaction points, discretizing free space via local grid expansion, enforcing minimum inter-node and node-edge distance constraints derived from robot dimensions, and applying transport demand-driven $K$-shortest path pruning. The method is evaluated across three intralogistics environments using two multi-agent pickup and delivery (MAPD) solvers against three baselines: a reaction-diffusion sampling method (GSRM), an 8-connected grid, and random sampling. Under Priority Inheritance with Backtracking (PIBT), the proposed method outperforms GSRM by $1.2$--$23.4\,\%$ at maximum fleet size, the grid by at least $9.1\,\%$, and random sampling by more than $10.4\,\%$ across all environments, with a space-time A$^*$ solver confirming these results. It further attains near-optimal normalized path lengths of $1.03$--$1.05$ and the highest inter-station connectivity at comparable roadmap complexity.
\end{abstract}


\section{Introduction}

The growing complexity of modern intralogistics has intensified interest in
mobile robot systems for material transport~\cite{zuzek_simulation-based_2023}.
Such systems can enhance transport efficiency, reduce costs, and
adapt flexibly to environmental changes and fluctuating transport
demand~\cite{sabattini_technological_2013}.
Achieving high system throughput and preventing congestion and deadlocks depends
not only on the fleet management system but also fundamentally on the design of
the underlying roadmap~\cite{zuzek_simulation-based_2023}.
A \textit{roadmap} is a graph-based representation of the navigable space, where
nodes denote discrete waypoints and edges represent feasible straight-line
connections.
In this work, \textit{mobile robot} encompasses both automated guided
vehicles~(AGVs) and autonomous mobile robots~(AMRs) capable of following
virtual roadmaps.

Roadmap design involves a fundamental trade-off between redundancy and
complexity.
\textit{Redundancy}, defined as the number of alternative paths between
station interaction points, enables robust coordination by reducing the risk of congestion and
deadlocks, and is strongly linked to efficient free-space utilization.
\textit{Complexity}, determined by the number of nodes and edges and their
geometric relations, affects the computational expense of path planning 
and can cause severe performance degradation in large-scale
systems~\cite{henkel_gsrm_2024}.
A well-designed roadmap therefore maximizes redundancy with short path
connections while minimizing structural complexity.

Despite their importance, roadmaps in industrial practice are still
predominantly generated manually, a process that is time-consuming, costly, and
often suboptimal~\cite{sabattini_technological_2013, digani_automatic_2014,
petrovic_improving_2023}.
Automated roadmap generation is accordingly essential to enable fast,
data-driven adaptation to changing environments and transport demand.

\subsection*{Related Work}

Automated roadmap generation has been explored from several perspectives.
Classical single-robot path planning algorithms, such as the probabilistic
roadmap method~(PRM)~\cite{geraerts_comparative_2004}, visibility
graphs~\cite{asano_visibility-polygon_1985}, rapidly-exploring random
trees~(RRT)~\cite{lavalle_rapidly-exploring_1998}, and Voronoi
diagrams~\cite{wallgrun_autonomous_2005}, are foundational for navigating known environments in single-robot settings.
However, these methods were not designed to sustain concurrent multi-robot
operation, as they do not provide
mechanisms to guarantee redundant alternative paths for an entire fleet~\cite{digani_automatic_2014}.

To address the specific requirements of mobile robot fleets, one major line of research focuses on \textit{grid-based} discretization.
These methods divide the free space into an equidistant grid, upon which various automated roadmap generation strategies have been developed, including reinforcement learning~\cite{kozjek_reinforcement-learning-based_2021}, ant colony optimization~(ACO)~\cite{petrovic_improving_2023,
zuzek_simulation-based_2023, vrabic_bio-inspired_2025}, shortest-path
algorithms combined with fuzzy logic~\cite{uttendorf_fuzzy-enhanced_2015,
uttendorf_fuzzy_2016}, and rule-based approaches~\cite{bang_virtual_2023}.
Most of these methods reduce the full grid by pruning nodes and edges
irrelevant to the specified transport demand and by assigning directional edge constraints.
Notably, the method of~\cite{zuzek_simulation-based_2023} takes the full intralogistics problem, including layout, task distribution, fleet size, and dispatching policy, as input, allowing the required level of redundancy to be adaptively determined.

Grid-based methods are attractive for their conceptual simplicity, regular node spacing and geometrically compact structure, but they are fundamentally constrained by their spatial resolution.
This constraint forces node placement onto a fixed lattice, which prevents
high-fidelity modeling of real-world environments.
In environments with narrow or irregularly shaped corridors, fixed-resolution grids can produce poorly connected or even disconnected roadmaps.
Furthermore, the achievable path lengths between nodes are bounded below by the Manhattan distance, which can be significantly longer than the Euclidean shortest path.

A second major line of research focuses on \textit{continuous-space} roadmaps, which offer greater geometric flexibility by operating without a fixed spatial resolution.
Proposed discretization strategies include the medial axis transform followed by linear programming optimization~\cite{kleiner_armo_2011}, rule-based algorithms~\cite{digani_automatic_2014, stenzel_automated_2021,
stenzel_automated_2022}, iterative Voronoi diagram generation with imaginary obstacles~\cite{aryadi_redundant_2023}, random sampling combined with stochastic gradient descent and Delaunay triangulation~\cite{henkel_optimized_2020}, and a Gray-Scott reaction-diffusion system combined with Delaunay triangulation~\cite{henkel_gsrm_2024}.
The latter, named GSRM, generates query-efficient, well-connected and redundant roadmaps tailored to the environmental geometry and has been shown to outperform other node-sampling strategies on path planning efficiency.
In~\cite{digani_automatic_2014}, a rule-based method is presented that
explicitly accounts for robot dimensions within corridors and intersections, with the goal of maximizing coverage, connectivity, and redundancy. This work is extended in~\cite{stenzel_automated_2021} to more cluttered environments, and a broader evaluation framework based on multi-agent pathfinding and graph theory is introduced in~\cite{stenzel_automated_2022}.

Continuous-space methods offer high geometric fidelity, but two important
limitations are consistently observed.
First, practical minimum distance constraints between nodes and edges, required to ensure that a robot occupying one roadmap element does not physically obstruct another, are largely neglected~\cite{henkel_optimized_2020, henkel_gsrm_2024}. Without such constraints, robots moving along adjacent edges or occupying nearby nodes can mutually block each other, reducing effective throughput in dense fleets.
Second, none of the existing continuous-space methods incorporates station positions or specified transport demand into the generation process.
These methods aim to produce uniformly well-connected roadmaps with high spatial coverage, but they do not adapt the roadmap structure to the actual operational requirements of the system.

\subsection*{Contributions}

\begin{figure}[!t]
\centering
\includegraphics[width=3.3in]{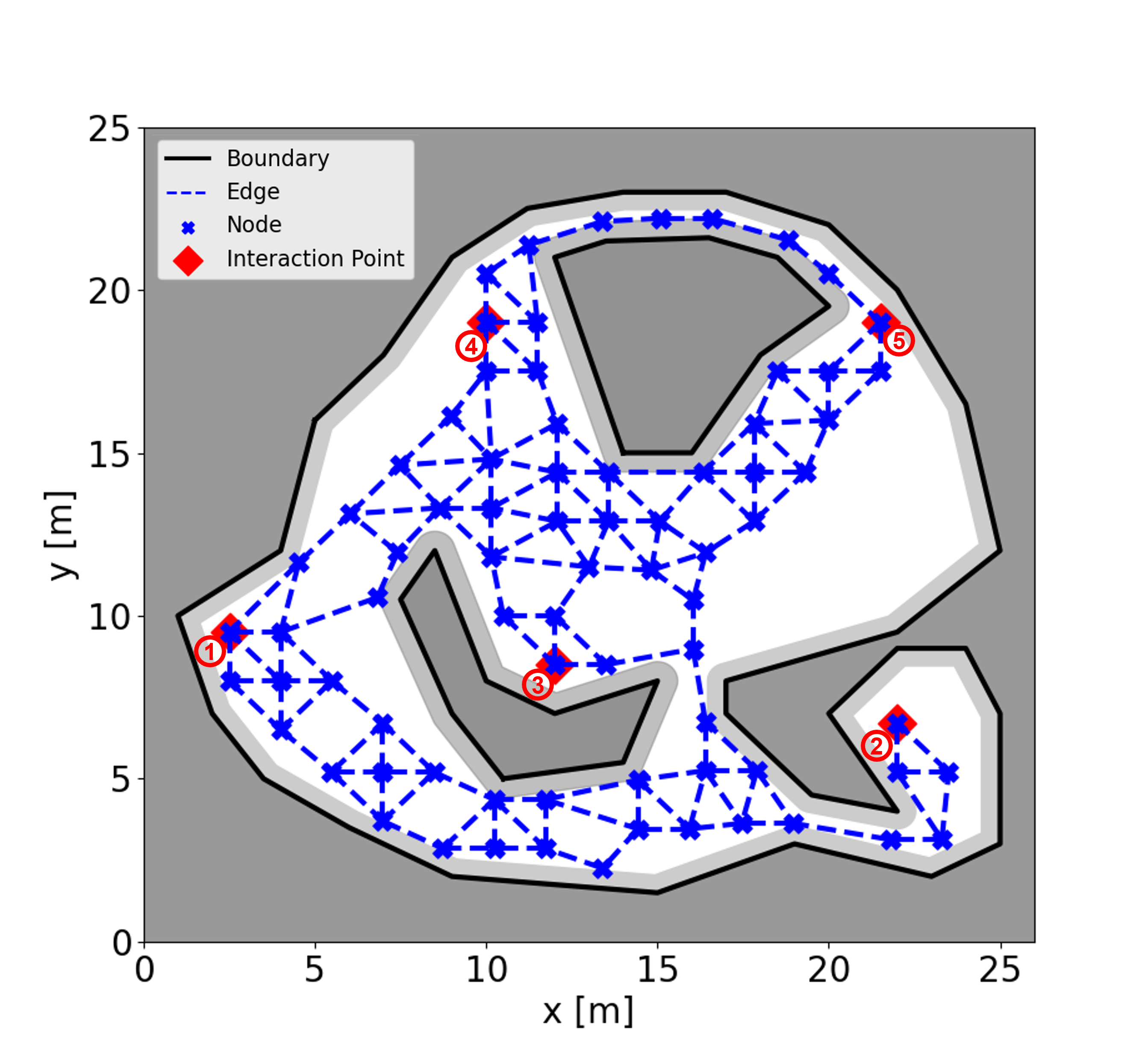}
\caption{Roadmap for an abstract environment with non-uniform transport demand (see Tab.~\ref{tab:transport_matrix}). Obstacles (dark gray) are geometrically
expanded (light gray) to define the free space. The roadmap provides redundant,
continuous-space paths (blue) adapted to both environmental geometry and
transport demand between interaction points.}
\label{fig_Roadmap_abstract}
\end{figure}

The existing literature reveals a critical gap: grid-based methods are
operationally robust but geometrically constrained, while continuous-space
methods offer geometric fidelity but neglect the distance constraints and
demand structure required for efficient fleet coordination.
This paper proposes a novel automated roadmap generation method that addresses
this gap.
Fig.~\ref{fig_Roadmap_abstract} visualizes a representative result for an abstract intralogistics environment with non-uniform transport demand (see Tab.~\ref{tab:transport_matrix}), demonstrating the method's ability to adapt to the environmental geometry while providing redundant, demand-tailored paths between station interaction points.

The main contributions of this paper are:

\begin{enumerate}
    \item A geometry-aware free-space discretization strategy incorporating
    station interaction points, convex corner points, and local grid
    expansion, generating continuous-space roadmaps with near-optimal
    path lengths and high inter-station redundancy. Transport demand-driven $K$-shortest path pruning reduces graph complexity while preserving routing redundancy.

    \item An explicit formulation and enforcement of minimum inter-node and
    node-edge distance constraints derived from mobile robot dimensions, ensuring neighboring roadmap elements can be occupied simultaneously without mutual interference, enabling efficient coordination in dense fleets.

    \item Demonstration that the proposed method outperforms
    established baselines, GSRM~\cite{henkel_gsrm_2024}, grid-based discretization, and random sampling, across three intralogistics environments of varying scale, achieving higher throughput in multi-agent pickup and delivery (MAPD) simulations as well as greater redundancy and shorter path lengths with comparable structural complexity.
\end{enumerate}

\section{Problem Formulation}
\label{sec:problem_formulation}

The proposed method takes as input a digital map of the intralogistics
environment, the physical dimensions of the mobile robots, and the system
transport demand, and produces a continuous-space roadmap tailored to these
inputs.
The environment is represented by polygonal objects in continuous
two-dimensional space, outer boundary, obstacles, and stations, enabling
high-fidelity modeling without the resolution constraints of a fixed
grid. While the outer boundary and obstacles are purely geometric, stations include one or more interaction points that serve as start and end points for transport tasks and may also have a geometric expansion acting as obstacle (compare Fig.\ref{fig_Roadmap_abstract} and Fig.\ref{fig_Maps}).
Map data may originate from CAD models, digital twins, or laser scanner data. Post-processing may be required to extract geometric features and convert environmental objects into polygonal representations~\cite{aryadi_redundant_2023,
beinschob_semi-automated_2017}.

\subsection{Free Space}

In the configuration space $C$, the state of a mobile robot is defined by its position $(x,y)$ and orientation $\theta$.
The free space is the set of all collision-free configurations:
\begin{equation}
\label{eq:C-Space}
C_{\text{free}} = C \setminus C_{\text{obs}},
\end{equation}
where $C_{\text{obs}}$ is the obstacle space.
To compute $C_{\text{free}}$, all geometric objects are expanded by
$r_{\text{rob}} + d_s$ (see Fig.~\ref{fig_Roadmap_abstract}), where $r_{\text{rob}}$ is the rotation radius of the robot and $d_s$ is a safety distance.
For robots with rotation-in-place kinematics (e.g., differential drive), the
footprint sweeps a circle of radius $r_{\text{rob}}$. Any center position
within $r_{\text{rob}} + d_s$ of an obstacle boundary is therefore infeasible.
The safety distance accounts for control imprecision, localization
inaccuracies, and permitted path deviations.

\subsection{Roadmap Graph and Distance Constraints}

\begin{figure}[!t]
\centering
\includegraphics[width=1.8in]{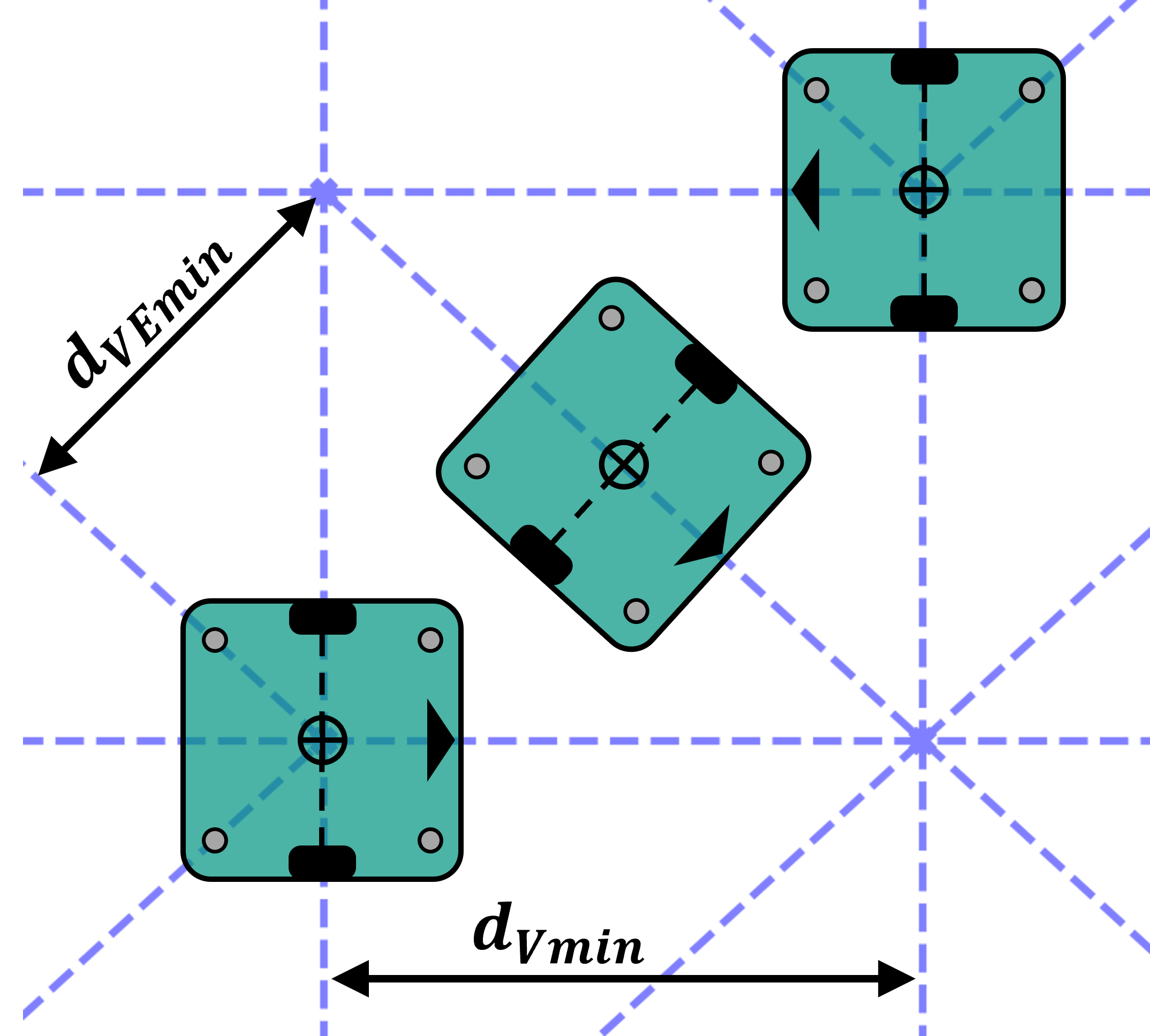}
\caption{Minimum distance constraints for roadmap nodes and edges.
$d_{V\text{min}}$: minimum inter-node distance;
$d_{VE\text{min}}$: minimum node-edge distance.}
\label{fig_Distance_Constraints}
\end{figure}

The roadmap is a planar graph $G = (V, E)$ with $V, E \subset C_{\text{free}}$,
where nodes $v \in V$ are discrete waypoints and edges $e = (v_i, v_j) \in E$
are bidirectional straight-line connections.
Planarity is required to avoid implicit intersection points at edge crossings,
which would create undefined robot interactions.
To enable simultaneous, non-interfering occupancy of adjacent elements by
different robots, two minimum distance constraints are imposed
(Fig.~\ref{fig_Distance_Constraints}):

\subsubsection{Minimum inter-node distance}
\begin{equation}
\label{eq:d-min-nodes}
\|v_i - v_j\| \;\geq\; d_{V\text{min}} = 2(r_{\text{rob}} + d_s),
\quad v_i \neq v_j,
\end{equation}
ensuring two robots at neighboring nodes do not overlap, as each occupies a
footprint of radius $r_{\text{rob}} + d_s$.

\subsubsection{Minimum node-edge distance}
\begin{equation}
\label{eq:d-min-nodes-edges}
d(v_k,\, e) \;\geq\; d_{VE\text{min}} = r_{\text{rob}} + \frac{w_{\text{rob}}}{2} + 2d_s,
\quad v_k \notin \{v_i, v_j\},
\end{equation}
where $w_{\text{rob}}$ is the robot width.
A robot traversing edge $e$ following the straight-line connection occupies a corridor of half-width
$\tfrac{w_{\text{rob}}}{2} + d_s$; a robot at node $v_k$ occupies a circle of
radius $r_{\text{rob}} + d_s$.
This constraint keeps these footprints separated, preventing interference
between a moving and a stationary robot.
The robots in this work are square-shaped with $r_{\text{rob}} = 0.5$~m,
$w_{\text{rob}} = 0.7$~m, and $d_s = 0.1$~m, yielding $d_{V\text{min}} = 1.2$~m and $d_{VE\text{min}} = 1.05$~m.

\subsection{Transport Demand}

The system transport demand is represented by a matrix
$T \in \mathbb{N}^{n \times n}$, where $n$ is the number of station interaction points. Each entry $T_{ij}$ specifies the expected number of transport tasks per time unit from interaction point $i$ to point $j$;
$T_{ij} = 0$ indicates no direct demand between that pair. Both loaded and empty return trips are encoded as separate entries.

Tab.~\ref{tab:transport_matrix} shows the matrix for the abstract environment of Fig.~\ref{fig_Roadmap_abstract}, with $n = 5$ interaction points. Station~1 acts as source and station~5 as sink; stations~2--4 represent intermediate process stations with ingoing and outgoing
material flow. Empty return trips are captured by $T_{51} = 5$. The sparse, asymmetric structure is representative of directed material flow in intralogistics. Matrices with higher demand values increase the number of retained $K$-shortest paths per station pair and thus roadmap redundancy.

\begin{table}[ht]
\centering
\caption{Transportation matrix of the abstract environment
(Fig.~\ref{fig_Roadmap_abstract}): number of transport tasks per time unit
between station interaction points.}
\label{tab:transport_matrix}
\begin{tabular}{c|cccccc}
\textbf{From\textbackslash To} & \textbf{1} & \textbf{2} & \textbf{3}
    & \textbf{4} & \textbf{5} \\
\hline
\textbf{1} & 0 & 4 & 0 & 1 & 0 \\
\textbf{2} & 0 & 0 & 4 & 2 & 0 \\
\textbf{3} & 0 & 0 & 0 & 2 & 0 \\
\textbf{4} & 0 & 0 & 0 & 0 & 5 \\
\textbf{5} & 5 & 0 & 0 & 0 & 0 \\
\end{tabular}
\end{table}

\section{Automated Roadmap Generation}
\label{sec:roadmap_generation}

\begin{figure*}[!t]
\centering
\subfloat[Visibility graph with convex corner nodes.]
    {\includegraphics[width=1.35in]{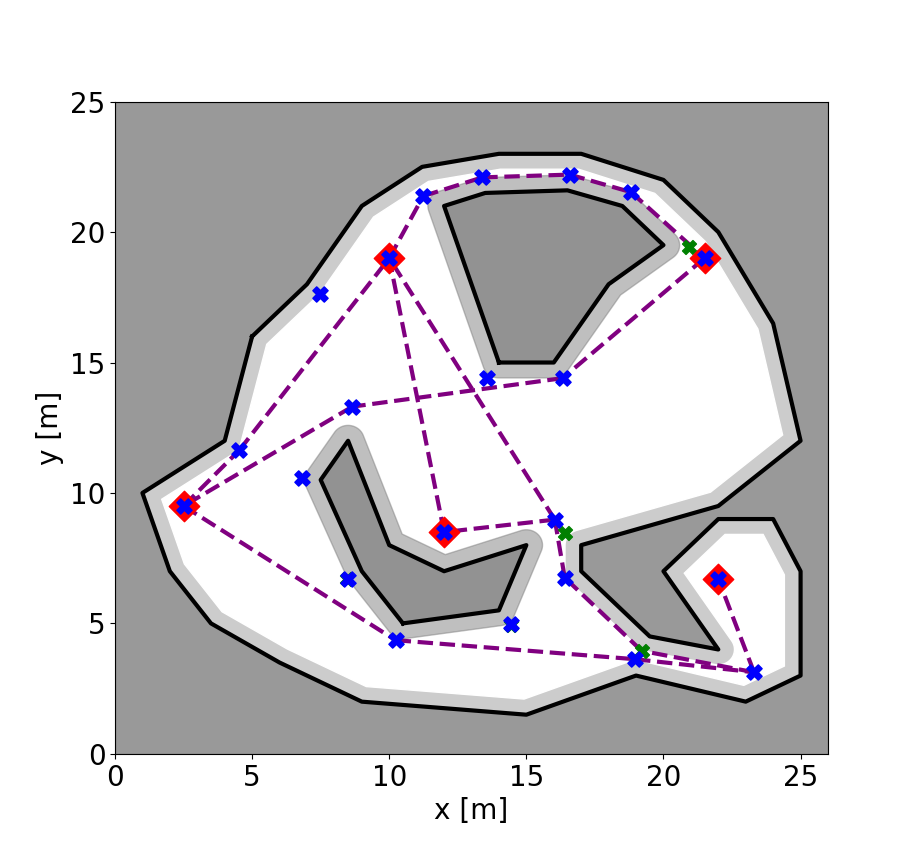}
    \label{fig_Visibility_Graph}}
\hfil
\subfloat[Discretized free space after local grid expansion.]
    {\includegraphics[width=1.35in]{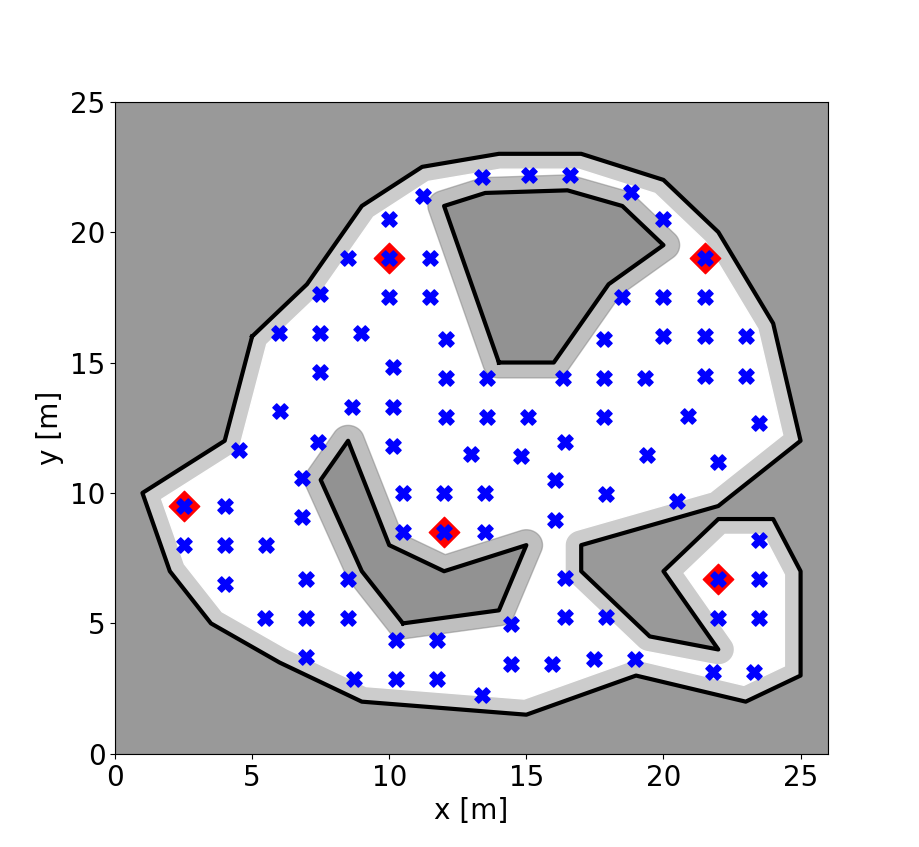}
    \label{fig:discr_roadmap}}
\hfil
\subfloat[Full roadmap after Delaunay triangulation.]
    {\includegraphics[width=1.35in]{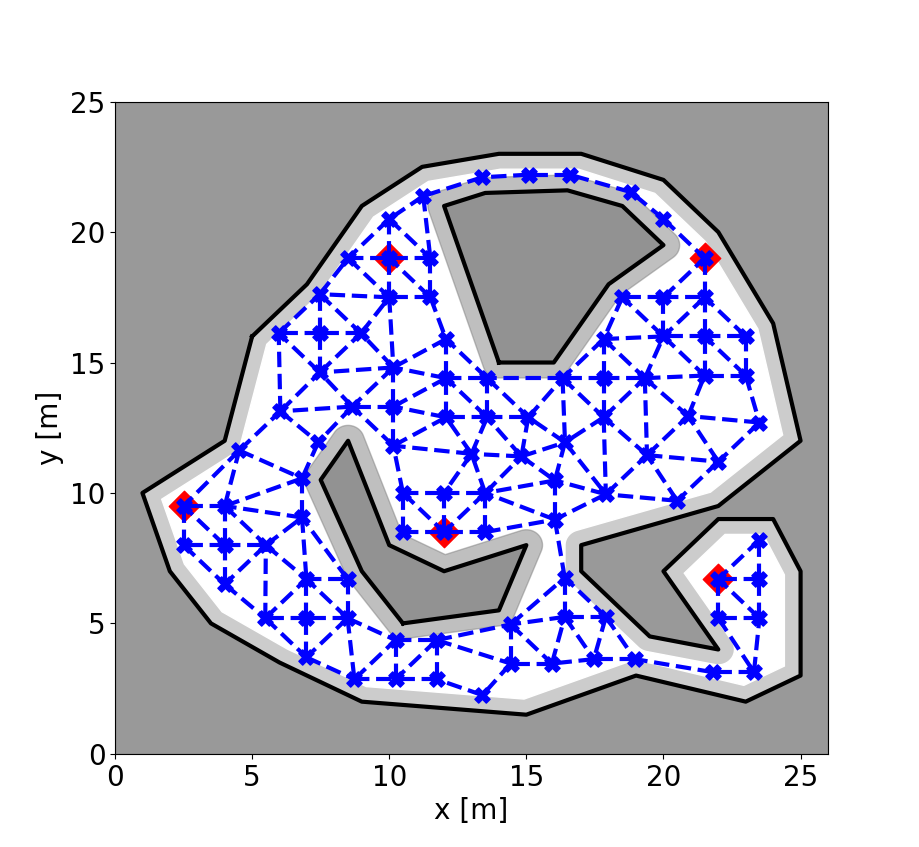}
    \label{fig:full_roadmap}}
\hfil
\subfloat[Reduced roadmap after $K$-shortest path pruning.]
    {\includegraphics[width=1.35in]{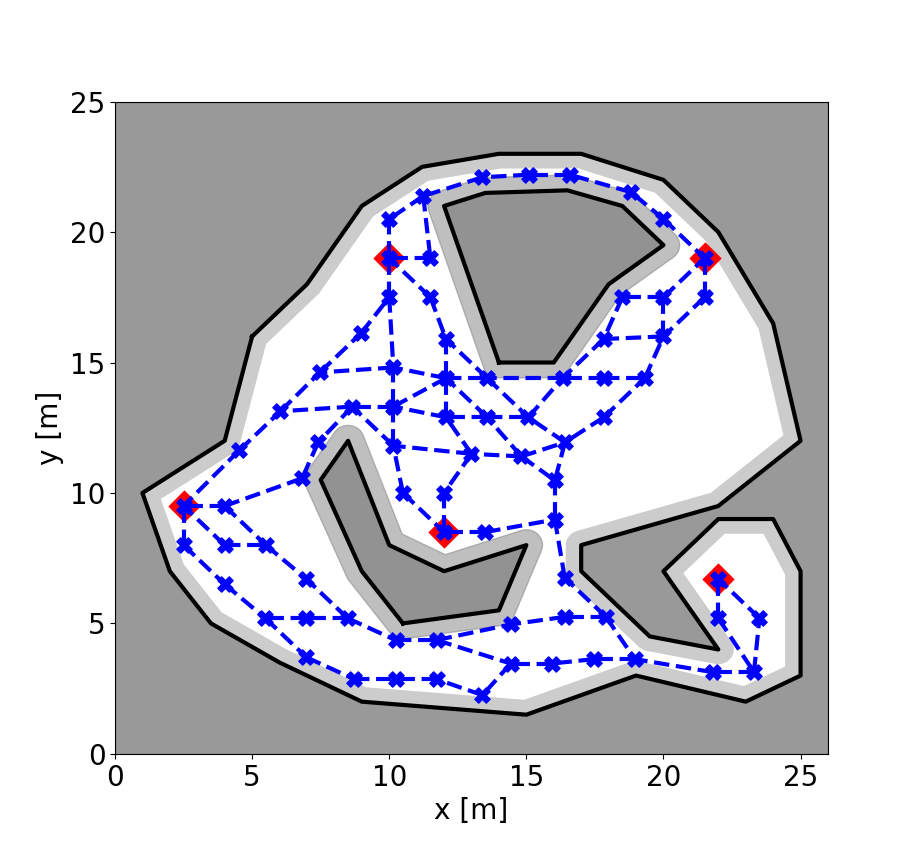}
    \label{fig:opt_roadmap}}
\hfil
\subfloat[Reduced roadmap after post-processing.]
    {\includegraphics[width=1.35in]{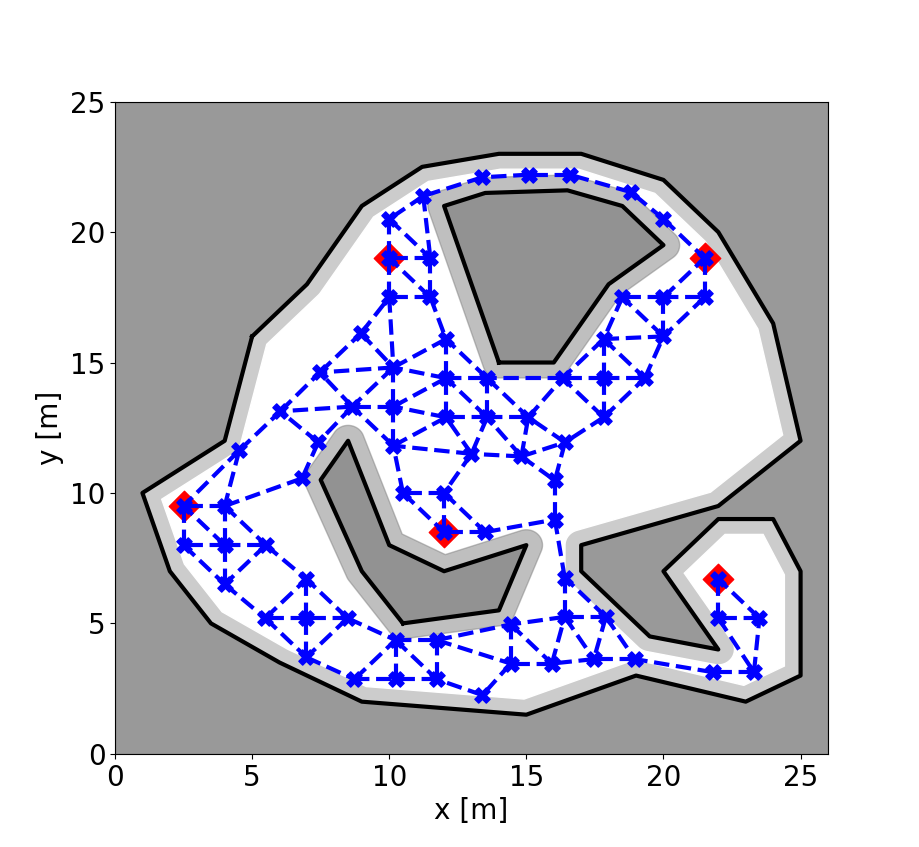}
    \label{fig:opt_roadmap_after}}
\caption{Automated roadmap generation process illustrated for an abstract environment.
(a)~Visibility graph connecting interaction points with transport demand; blue crosses: added nodes
$v \in V_{\text{st}} \cup V_{\text{co}}$; green crosses: excluded corner
points violating $d_{V\text{min}}$.
(b)~Node distribution after free space discretization using local grid expansion.
(c)~Full roadmap after edge construction using Delaunay triangulation.
(d)~Reduced roadmap after $K$-shortest path pruning based on transport demand from Tab.~\ref{tab:transport_matrix}, before post-processing.
(e)~Final roadmap after post-processing using second Delaunay triangulation run.}
\label{fig_automated_roadmap_generation}
\end{figure*}

The proposed method generates a roadmap $G = (V, E)$ in three main steps:
(i)~free-space discretization, (ii)~edge construction, and
(iii)~transport demand-driven pruning, illustrated in
Fig.~\ref{fig_automated_roadmap_generation}.
A pre-processing stage first derives $C_{\text{free}}$ from the polygonal
environment model as described in Sec.~\ref{sec:problem_formulation}.
The following description is illustrated using the abstract environment of Fig.~\ref{fig_Roadmap_abstract}. The three intralogistics environments used for quantitative evaluation are presented in Sec.~\ref{sec:evaluation_methodology}.

\subsection{Free-Space Discretization}
\label{subsec:discretization}

The discretization pursues three objectives: (i)~adapt to arbitrarily shaped
environments, (ii)~ensure efficient free-space utilization, and (iii)~enable
short path connections between station interaction points.

\subsubsection{Interaction points and corner nodes}
Transport tasks start and end at interaction points, so nodes are
first placed at all interaction point positions.
Let $V_{\text{st}} \subset C_{\text{free}}$ denote this set.
Next, convex corner points of $C_{\text{free}}$ are identified: since the
visibility graph, encoding Euclidean shortest paths between interaction
points~\cite{lavalle_planning_2006}, consists almost entirely of convex corner
points, placing nodes there directly supports short path connections between interaction points.
A corner point is added to $V_{\text{co}}$ only if Eq.~\eqref{eq:d-min-nodes}
is satisfied with respect to all previously added nodes. When candidates
compete for the same region, priority is assigned by participation count in
visibility-graph shortest paths weighted by $T_{ij}$, see Fig.~\ref{fig_Visibility_Graph}.

\subsubsection{Local grid expansion}
To achieve dense coverage of $C_{\text{free}}$ beyond $V_{\text{st}} \cup
V_{\text{co}}$, local grids are grown iteratively from all existing nodes.
For a node at $(x, y)$ with grid spacing $d_g$, the 8-connected neighborhood nodes $v'$ are:
\begin{align}
%\label{eq:local_grid}
%v' \in \bigl\{(x + i \cdot d_g,\; y + j \cdot %d_g) \;\big|\;
%    i, j \in \{-1, 0, 1\},\; (i,j) \neq (0,0) \bigr\}.
%\end{align}
%\begin{align}
v' \in \{(x + i \cdot d_g,\; y + j \cdot d_g) \mid\; & i, j \in \{-1, 0, 1\}, \notag \\
                                                  & (i, j) \neq (0,0)\}.
\end{align}
Starting from hop radius~1, the grid expands iteratively by incrementing the maximum value of grid hops $i$ and $j$, with interaction point neighborhoods expanded first.
A candidate is added only if it lies within $C_{\text{free}}$ and satisfies
Eq.~\eqref{eq:d-min-nodes}.
The grid spacing $d_g$ is derived by requiring that, within a regular local grid, diagonal edges of length $d_g\sqrt{2}$ satisfy Eq.~\eqref{eq:d-min-nodes-edges}: the perpendicular
distance from any collinear neighbor to a diagonal edge equals $d_g/\sqrt{2}$, giving:
\begin{equation}
\label{eq:d_g}
d_g \;\geq\; \sqrt{2}\,d_{VE\text{min}}
    = \sqrt{2}\Bigl(r_{\text{rob}} + \tfrac{w_{\text{rob}}}{2} + 2d_s\Bigr),
\end{equation}
which yields $d_g \geq 1.485$~m for the robot parameters of
Sec.~\ref{sec:problem_formulation}.
The resulting node distribution is shown in Fig.~\ref{fig:discr_roadmap}.

\subsection{Edge Construction}
\label{subsec:edges}

Edges are constructed using Delaunay triangulation~\cite{delaunay_sur_1934},
which connects nodes by non-overlapping triangles while maximizing the minimum
interior angle, ensuring planar graphs while promoting path geometry without sharp turns.
An edge $e = (v_i, v_j)$ is added to $G$ if:
\begin{enumerate}[]
    \item $e \subset C_{\text{free}}$: the edge lies entirely within free space, and
    \item $d(v_k, e) \geq d_{VE\text{min}}$ for all $v_k \in V \setminus \{v_i, v_j\}$: the node-edge distance constraint of Eq.~\eqref{eq:d-min-nodes-edges}
    is satisfied for all non-endpoint nodes.
\end{enumerate}
The resulting full roadmap is shown in Fig.~\ref{fig:full_roadmap}.
All edges are bidirectional; the assignment of directional constraints is
deferred to the fleet management system, which can dynamically impose them
based on traffic policies or real-time conditions.

\subsection{Transport Demand-Driven Pruning}
\label{subsec:roadmap_pruning}

The full roadmap is pruned to retain only nodes and edges contributing to the
specified transport demand, pursuing three objectives: (i)~maintain
inter-station redundancy for high throughput, (ii)~reduce graph complexity, and
(iii)~make space available for other operational processes. The effect of pruning on graph complexity and throughput is quantified in Sec.~\ref{subsec:pruning_analysis}.

For each ordered pair $(i,j)$ with $T_{ij} > 0$, a set of $k_{ij}$
loop-free paths is computed using an extended Yen's $K$-shortest path
algorithm~\cite{yen_finding_1971}, with $k_{ij} = \lceil T_{ij} /
T_{\text{unit}} \rceil$. $T_{\text{unit}}$ is a configurable granularity
parameter controlling the number of retained paths per station pair:
higher values reduce roadmap complexity at the cost of lower redundancy,
while $T_{\text{unit}} = 1$ yields the maximum demand-driven redundancy
and is used throughout this work. Since identical pruning parameters
are applied to all compared discretization methods, $T_{\text{unit}}$
acts as a controlled experimental variable.

To promote path diversity, edge costs are penalized after each found
path: the cost of every $e \in E_p$ is scaled by a factor
$\gamma = \alpha^{u_{ij}(e)}$, where $u_{ij}(e)$ counts prior traversals
of $e$ for pair $(i,j)$, biasing subsequent searches toward spatially
distinct paths. The penalty base $\alpha = 1.1$ is empirically selected
to balance path diversity and compactness. After all $k_{ij}$ paths for
a pair are found, $u_{ij}(e)$ of all edges is reset to~$0$.

Let $V_{\text{ksp}} = \bigcup_{i,j} V_p$ and $E_{\text{ksp}} =
\bigcup_{i,j} E_p$ denote the union of all retained nodes and edges.
All unused elements $V \setminus V_{\text{ksp}}$ and $E \setminus
E_{\text{ksp}}$ are removed (Fig.~\ref{fig:opt_roadmap}). Delaunay
triangulation is then applied a second time over $V_{\text{ksp}}$ as a
post-processing step to restore routing flexibility, adding edges subject
to conditions~(1) and~(2) of Sec.~\ref{subsec:edges} while preserving
a planar graph, with $d_{VE\text{min}}$ re-verified for all newly
inserted edges. The final roadmap is shown in
Fig.~\ref{fig:opt_roadmap_after}.

\section{Evaluation Methodology}
\label{sec:evaluation_methodology}

\begin{figure*}[!t]
\centering
\subfloat[Env.~1.]
    {\includegraphics[height=1.8in]{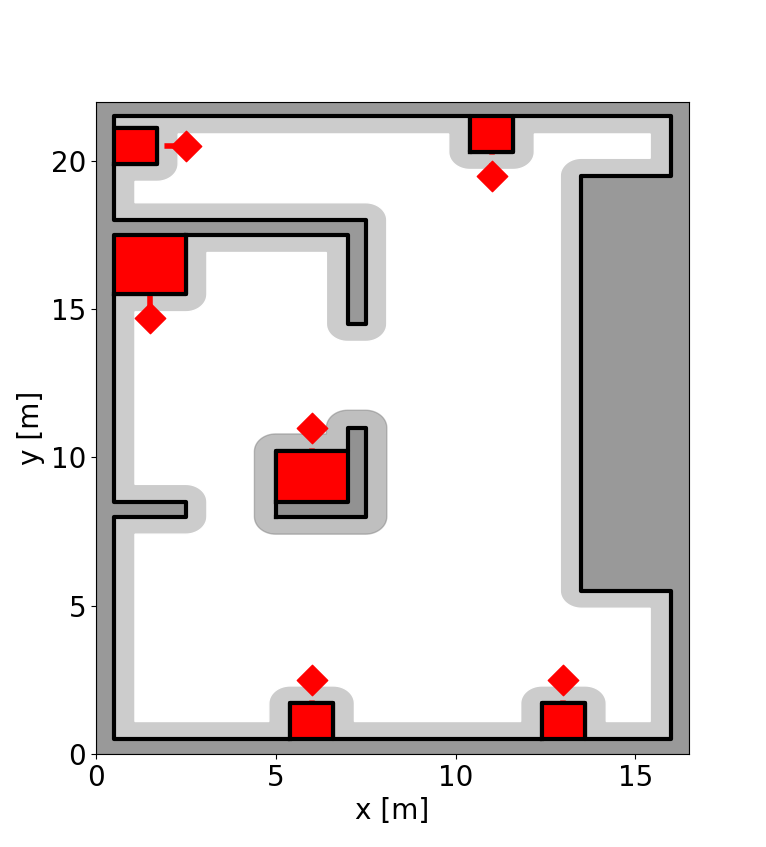}\label{fig:Map1}}
\hfil
\subfloat[Env.~2.]
    {\includegraphics[height=1.8in]{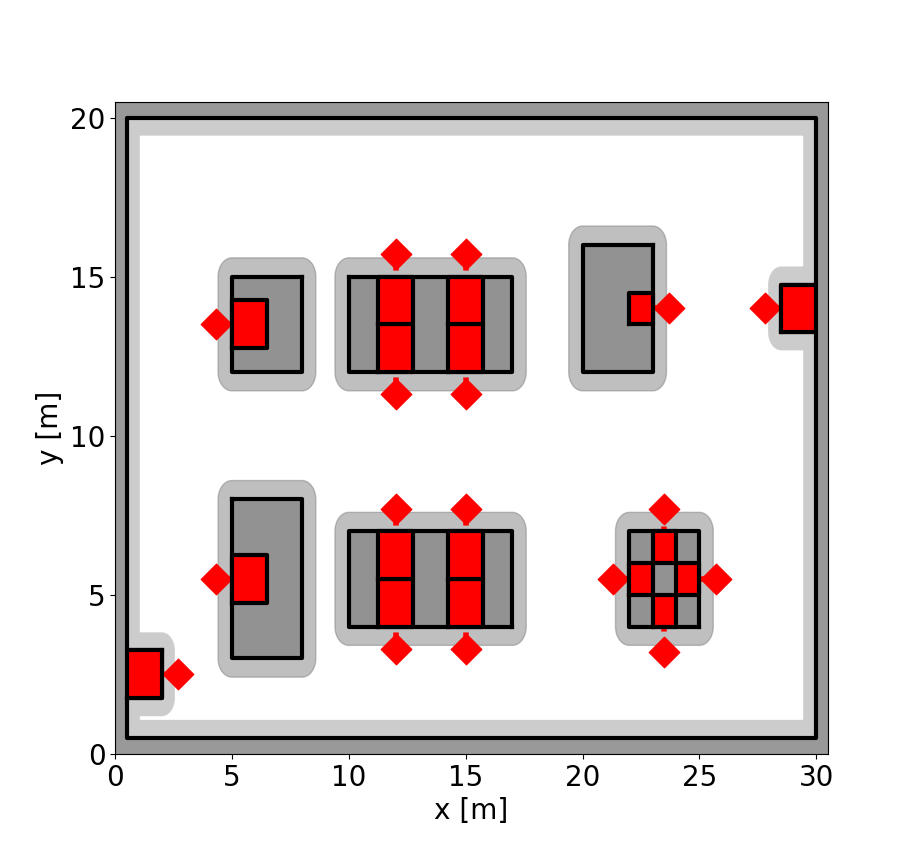}\label{fig:Map2}}
\hfil
\subfloat[Env.~3.]
    {\includegraphics[height=1.8in]{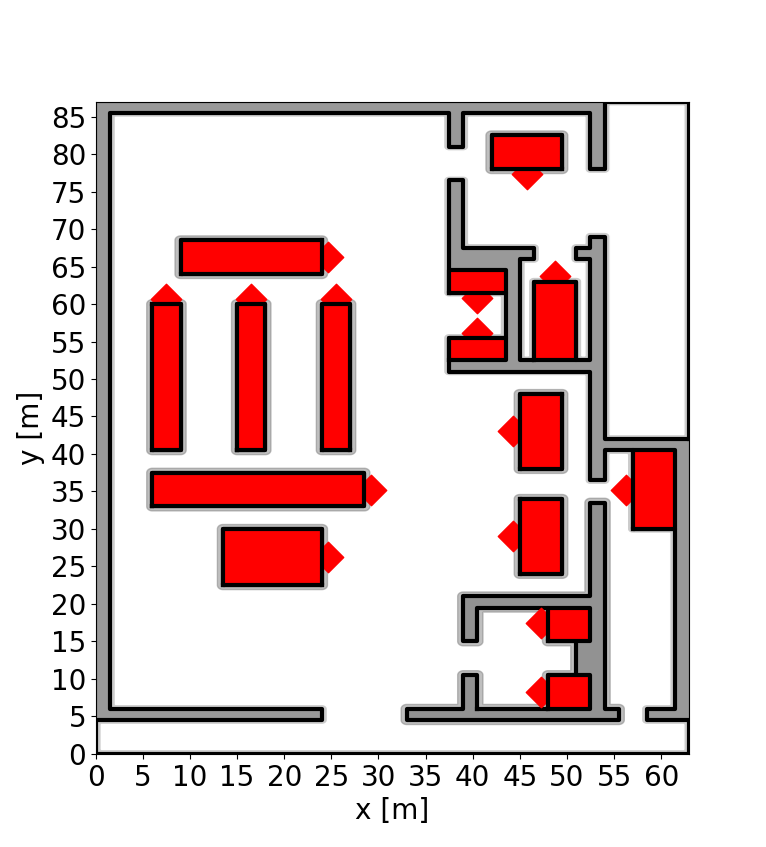}\label{fig:Map3}}
\caption{Intralogistics environments used for evaluation, shown without roadmap
overlay. Red areas are stations with interaction points marked by red rhombuses.}
\label{fig_Maps}
\end{figure*}

The proposed method is evaluated across three intralogistics environments of
varying scale using MAPD simulation-based throughput benchmarking as the
primary evaluation and graph-theoretical metrics as a complementary structural
analysis. For each environment, roadmaps are generated by the proposed method
and three baseline strategies, all subjected to identical edge construction,
pruning, and post-processing (Sec.~\ref{subsec:roadmap_pruning}), ensuring
that performance differences are attributable to the free-space discretization
strategy alone.

\subsection{Environments and Baselines}
\label{subsec:environments}

Three intralogistics environments from the literature are used, rescaled to
match the robot dimensions of Sec.~\ref{sec:problem_formulation} and shown
in Fig.~\ref{fig_Maps}.

\begin{itemize}
    \item \textbf{Env.~1}~\cite{uttendorf_fuzzy-enhanced_2015}:
    small-scale room-like layout with uniform transport demand between all
    station pairs ($T_{ij} = 3$).
    \item \textbf{Env.~2}~\cite{kozjek_reinforcement-learning-based_2021}:
    medium-scale matrix production system with directional material flow from
    the bottom-left to the upper-right station and empty return trips
    ($T$ comparable to Tab.~\ref{tab:transport_matrix}).
    \item \textbf{Env.~3}~\cite{uttendorf_fuzzy_2016,
    zuzek_simulation-based_2023}: large-scale layout with open space and
    partly narrow corridors, with uniform transport demand between all station
    pairs ($T_{ij} = 5$).
\end{itemize}

Three baseline discretization strategies are evaluated, with identical edge
construction and pruning across all methods.

\begin{itemize}
    \item \textbf{GSRM}~\cite{henkel_gsrm_2024}: nodes generated by a
    Gray-Scott reaction-diffusion system, reimplemented and integrated into
    the proposed framework. The diffusion resolution is adjusted so that
    generated nodes satisfy Eq.~\eqref{eq:d-min-nodes}, and interaction
    point nodes are placed first and given priority.
    \item \textbf{8-connected Grid}: nodes placed on a regular lattice with
    spacing $d_g$ per Eq.~\eqref{eq:d_g}, ensuring that diagonal connections
    satisfy Eq.~\eqref{eq:d-min-nodes-edges}. The lattice origin is set at
    the coordinate origin.
    \item \textbf{Random Sampling}: nodes placed uniformly at random within
    $C_{\text{free}}$, subject to Eq.~\eqref{eq:d-min-nodes}. The
    best-performing roadmap of ten independent runs, measured by throughput,
    is reported, providing a favorable estimate of random sampling performance
    and ensuring that conclusions are not artifacts of a poor random seed.
\end{itemize}

\begin{figure*}[!t]
\centering
\subfloat[Own.]
    {\includegraphics[width=1.6in]{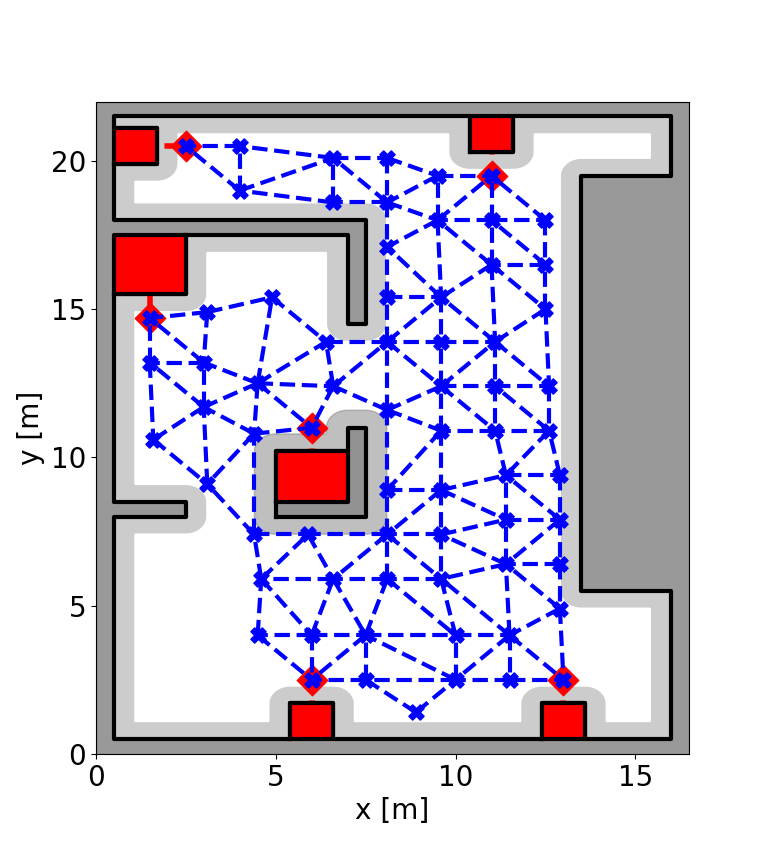}
    \label{fig:approach}}
\hfil
\subfloat[GSRM~\cite{henkel_gsrm_2024}.]
    {\includegraphics[width=1.6in]{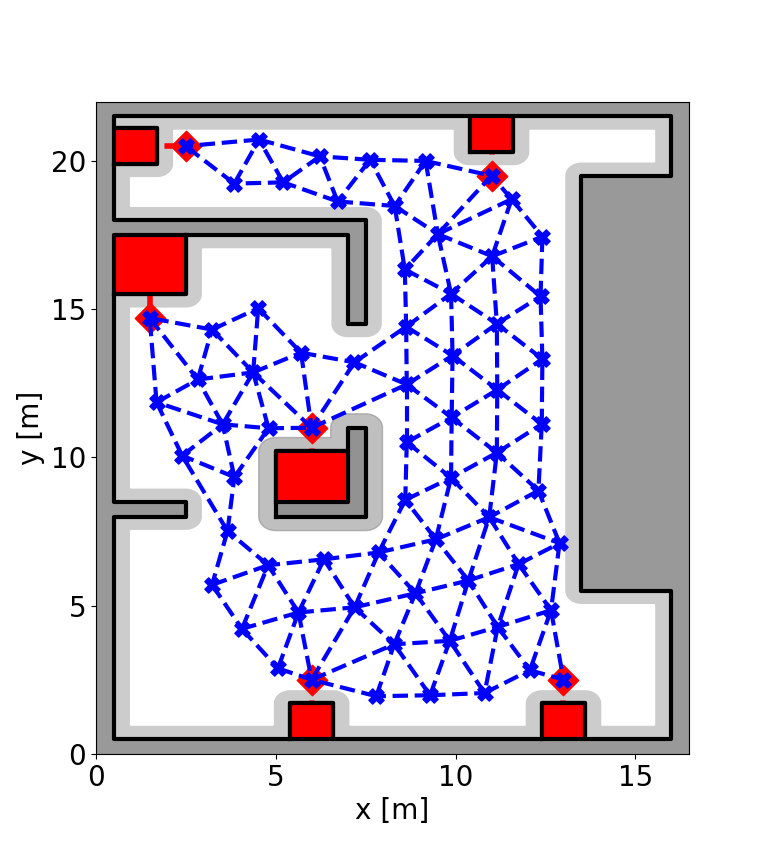}
    \label{fig:gsrm}}
\hfil
\subfloat[8-connected grid.]
    {\includegraphics[width=1.6in]{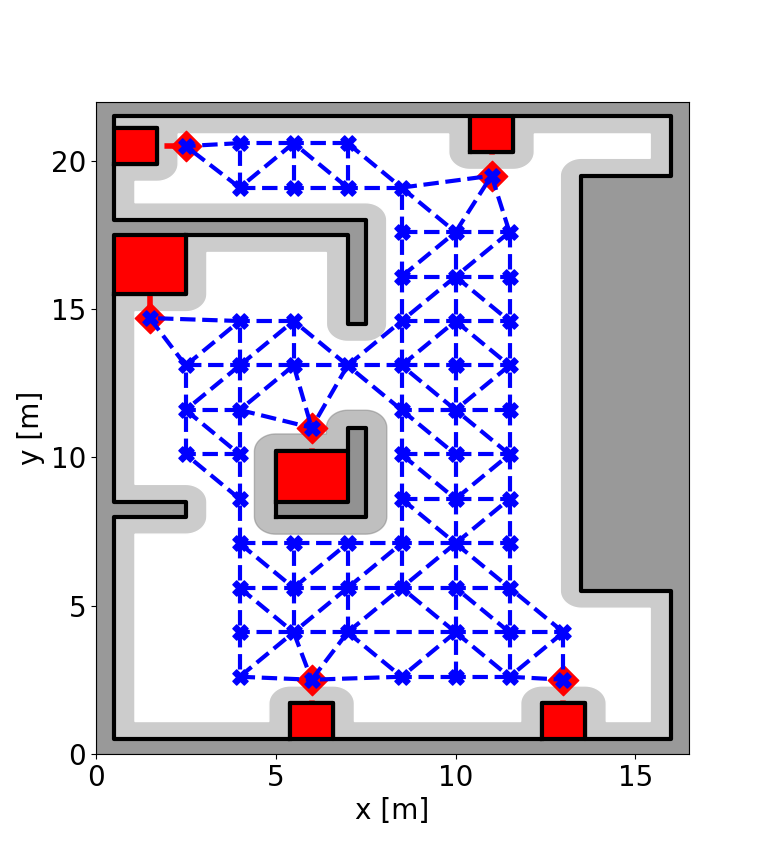}
    \label{fig:8-grid}}
\hfil
\subfloat[Random sampling.]
    {\includegraphics[width=1.6in]{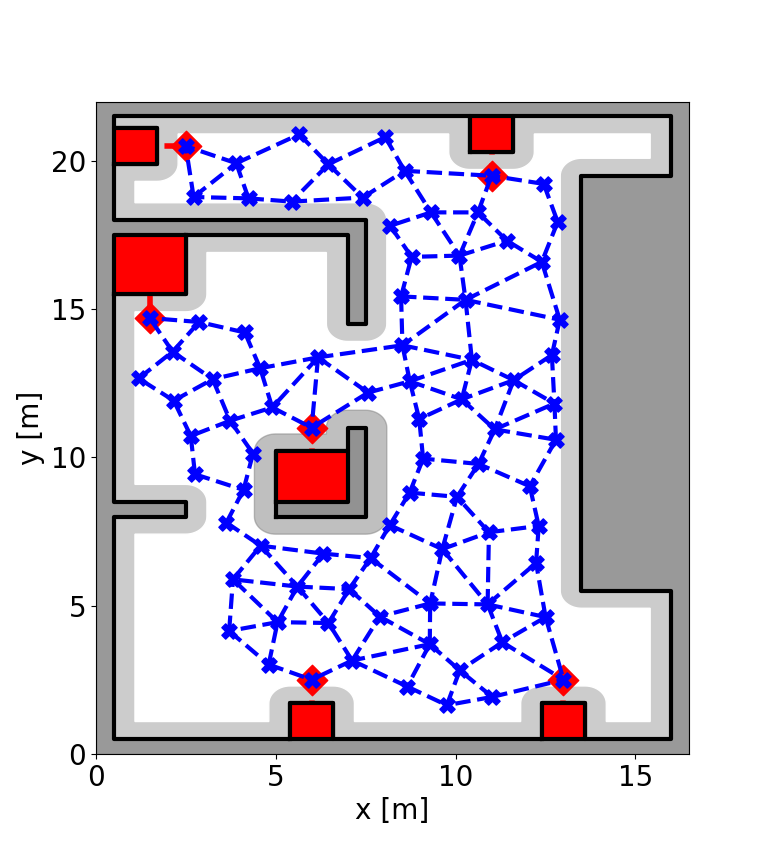}
    \label{fig:random}}
\caption{Roadmaps generated for Env.~1 by the proposed method and the
three baselines after identical edge construction and demand-aware pruning.}
\label{fig_Roadmaps}
\end{figure*}

Roadmaps for Env.~1 are compared qualitatively in
Fig.~\ref{fig_Roadmaps}.

\subsection{Multi-Agent Throughput Evaluation}
\label{subsec:simulation}

Operational performance is evaluated using two independent MAPD solvers,
providing evidence that roadmap quality rankings are consistent across
planning paradigms.

\subsubsection{Solvers}

The first solver is Priority Inheritance with
Backtracking~(PIBT)~\cite{okumura_priority_2022}, extended to directed
graphs and MAPD; the implementation is publicly
available\footnote{https://github.com/marvin-rdt/pypibt-mapd}.
PIBT is a one-step planner that resolves conflicts at each timestep via
priority inheritance without computing full agent trajectories in advance.
It scales well to high agent densities.

The second solver is a space-time A$^*$ planner for MAPD, based on
continuous-time space-time search with time-window node reservations,
inspired by Cooperative~A$^*$~\cite{silver_cooperative_2005} and
SIPP~\cite{phillips_sipp_2011}, and implemented in a discrete-event
simulation framework~(SimPy). Agents are planned sequentially using a
shared reservation table over node-time intervals, so that each new path
is conflict-free by construction. Station nodes permit concurrent
occupancy to simplify fleet management while all other nodes
are exclusive. In practice, most of the time stations 
are occupied by at most one agent. 
Simultaneous occupancy by two or three agents
occurs only briefly, primarily in Environment~1 at
larger fleet sizes. Due to its sequential
full-path planning, the space-time A$^*$ planner is evaluated at smaller
fleet sizes than PIBT.

Both solvers use identical roadmap inputs in Layout Interchange
Format~(LIF) with metric node coordinates, directed edges, and station
definitions. Tasks are dispatched in global FIFO order from a
pre-generated queue sampled from station pairs weighted by~$T$. Both
solvers use a graph-based node-occupancy abstraction, ensuring that
observed performance differences are attributable to roadmap structure.

\subsubsection{Time Model}

PIBT resolves conflicts in discrete unit timesteps. Goal selection uses
a shortest-path distance table computed with Euclidean edge lengths~$d_e$
via Dijkstra's algorithm. Physical time per agent is recovered as
$t_{\text{move}}=d_{\text{total}}/v$ and
$t_{\text{wait}}=n_{\text{wait}}\cdot\bar{d}_e/v$,
where $v=1.0$~m/s, $d_{\text{total}}$ is the total distance traversed,
$n_{\text{wait}}$ is the number of waiting timesteps, and $\bar{d}_e$
is the fleet-average edge length. The proposed method exhibits the
longest average edges across all environments ($1.84$--$1.92$~m,
vs.\ $1.63$--$1.84$~m for the baselines), so waiting steps 
incur a proportionally larger physical time penalty, 
making the throughput comparison conservative.

\subsubsection{Metrics and Experimental Protocol}

The primary metric is steady-state throughput:
$\tau = N_{\text{tasks}} / t_{\text{makespan}}$~[tasks/s],
where $t_{\text{makespan}} = \max_i(t_{\text{move},i} + t_{\text{wait},i})$.
Both solvers discard an equal number of warm-up and cool-down task
completions per run, isolating a steady-state window in which all agents
remain active. For PIBT, 300 tasks are excluded at each end of
$N_{\text{total}}^{\text{PIBT}} = 1600$, yielding
$N_{\text{tasks}}^{\text{PIBT}} = 1000$ evaluated tasks; for the
space-time A$^*$ planner, 30 tasks are excluded at each end of
$N_{\text{total}} = 260$, yielding $N_{\text{tasks}} = 200$ evaluated
tasks. Each configuration is repeated across 10 random seeds, with fleet
size ranges adapted to each environment and solver. Results are reported
as the median~$\pm$~IQR/2 to limit outlier influence.

\subsection{Graph-Theoretical Metrics}
\label{subsec:graph_metrics}

Three structural metrics characterize roadmap properties independently of
any fleet management system, following~\cite{digani_automatic_2014,
stenzel_automated_2021, stenzel_automated_2022}, and are reported in a
single compact table.

\begin{itemize}
    \item \textbf{Number of nodes $|V|$ and edges $|E|$}: structural
    complexity, determining the memory footprint and search space of
    path planning algorithms.
    \item \textbf{Mean node and edge connectivity}: the mean, over all
    pairs $(i,j)$ with $T_{ij} > 0$, of the minimum number of nodes
    (resp.\ edges) whose removal disconnects the pair, quantifying
    redundancy and fault tolerance~\cite{stenzel_automated_2022}.
    \item \textbf{Normalized mean shortest path length}: the mean ratio
    of the roadmap shortest path to the visibility graph shortest path
    distance over all pairs with $T_{ij} > 0$, computed via Dijkstra's
    algorithm. A value of~1 indicates a geometrically optimal path;
    values above~1 quantify the detour imposed by the discrete roadmap
    topology.
\end{itemize}

\section{Results and Evaluation}
\label{sec:evaluation_results}

\begin{figure*}[!t]
\centering
    {\includegraphics[width=\textwidth]{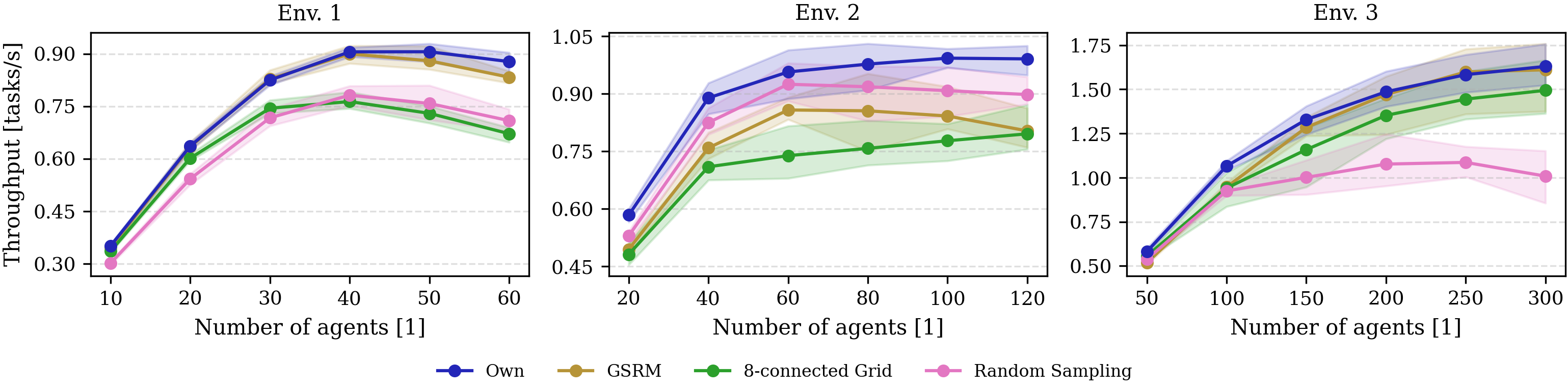}}
\caption{Steady-state throughput~$\tau$ of the PIBT solver as a function of
fleet size. Lines: median; bands: IQR across 10 seeds.}
\label{fig:results_throughput_pibt}
\end{figure*}

\begin{figure*}[!t]
\centering
    {\includegraphics[width=\textwidth]{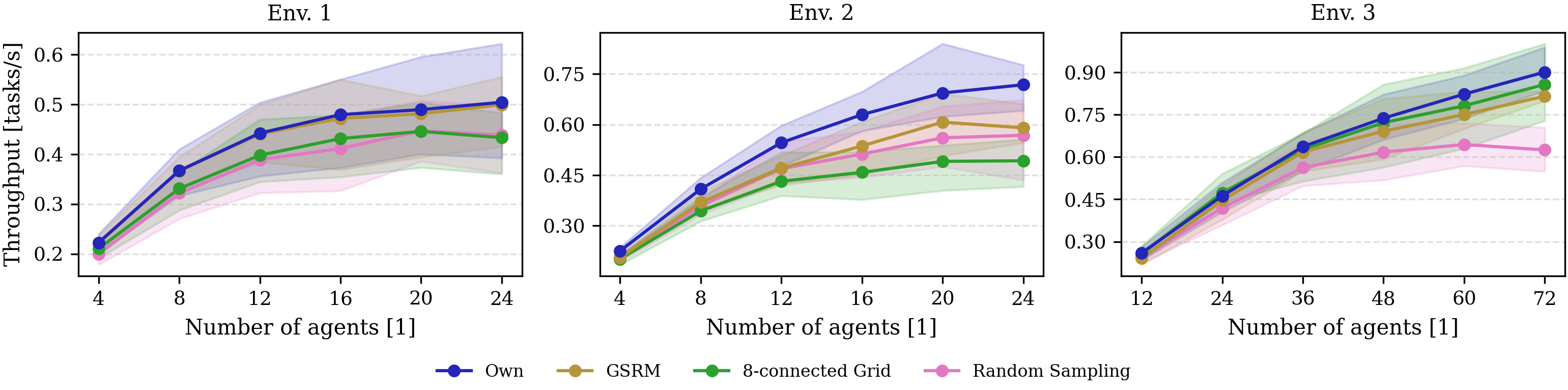}}
\caption{Steady-state throughput~$\tau$ of the space-time A$^*$ solver as a
function of fleet size. Lines: median; bands: IQR across 10 seeds.}
\label{fig:results_throughput_sta}
\end{figure*}

Figs.~\ref{fig:results_throughput_pibt}
and~\ref{fig:results_throughput_sta} present the throughput results for
all three environments under the PIBT and space-time A$^*$ solvers,
respectively. Tab.~\ref{tab:graph_metrics} summarizes the
graph-theoretical metrics. Since all methods share identical edge
construction, pruning, and post-processing, performance differences in
Sec.~\ref{subsec:simulation} and~\ref{subsec:graph_metrics} are
attributable to the free-space discretization strategy alone.
Sec.~\ref{subsec:pruning_analysis} separately quantifies the effect
of pruning on graph complexity and throughput.

\subsection{MAPD Simulations}
\label{subsec:simulation}
\textbf{Env.~1.}
The proposed method and GSRM substantially outperform the grid and random
sampling across all fleet sizes. Under PIBT, the proposed method peaks at
$0.906$~tasks/s ($|A|=50$), compared to $0.900$, $0.764$, and
$0.782$~tasks/s for GSRM, the grid, and random sampling ($|A|=40$). At
maximum fleet size ($|A|=60$), the proposed method yields $0.878$~tasks/s,
exceeding GSRM by $5.4\,\%$, the grid by $30.7\,\%$, and random sampling
by $23.8\,\%$. The space-time A$^*$ solver confirms the ranking: at
$|A|=24$ the proposed method reaches $0.504$~tasks/s, exceeding GSRM by
$1.0\,\%$, the grid by $16.4\,\%$, and random sampling by $15.0\,\%$.

\textbf{Env.~2.}
The proposed method achieves the highest throughput across all fleet sizes
under both solvers. Under PIBT, it peaks at $0.993$~tasks/s ($|A|=100$).
GSRM and random sampling peak
at $|A|=60$ ($0.858$ and $0.925$~tasks/s) and decline thereafter; the
grid increases gradually to $0.796$~tasks/s. At $|A|=120$, the proposed
method yields $0.991$~tasks/s, exceeding random sampling by $10.4\,\%$,
GSRM by $23.4\,\%$, and the grid by $24.6\,\%$. The sustained scaling is
attributable to the higher inter-station connectivity
(Tab.~\ref{tab:graph_metrics}) and geometry-aware node placement in narrow
corridors, which reduces conflict-induced wait times at high fleet
densities. The space-time A$^*$ solver corroborates this result: at
$|A|=24$, the proposed method reaches $0.718$~tasks/s, exceeding GSRM by
$21.7\,\%$, random sampling by $26.4\,\%$, and the grid by $45.8\,\%$.

\begin{table*}[ht]
\centering
\caption{Graph-theoretical metrics for the proposed method and three
baselines across the three environments. Green cells indicate the
best value per metric per environment; the ideal direction is given in
the rightmost column.}
\label{tab:graph_metrics}
\begin{tabular}{|l|cccc|cccc|cccc|c|}
\hline
\textbf{Metric} &
\multicolumn{4}{c|}{\textbf{Env.~1}} &
\multicolumn{4}{c|}{\textbf{Env.~2}} &
\multicolumn{4}{c|}{\textbf{Env.~3}} &
\textbf{Ideal} \\
\hline
& \textbf{Own} & \textbf{GSRM} & \textbf{Grid} & \textbf{Rand.} &
  \textbf{Own} & \textbf{GSRM} & \textbf{Grid} & \textbf{Rand.} &
  \textbf{Own} & \textbf{GSRM} & \textbf{Grid} & \textbf{Rand.} & \\
\hline
\textbf{Nodes $|V|$} &
69 & 68 & 67 & 74 &
135 & 135 & 129 & 127 &
728 & 799 & 747 & 722 & - \\
\textbf{Edges $|E|$} &
160 & 157 & 149 & 149 &
306 & 262 & 277 & 255 &
1788 & 1643 & 1855 & 1628 & - \\
\hline
\textbf{Mean node con.} &
\cellcolor{bestgreen}2.73 & 2.40 & 1.73 & 1.87 &
\cellcolor{bestgreen}3.86 & 2.39 & 2.89 & 2.82 &
\cellcolor{bestgreen}3.09 & 1.97 & 2.65 & 2.02 & max \\
\textbf{Mean edge con.} &
\cellcolor{bestgreen}2.73 & 2.53 & 2.20 & 2.20 &
\cellcolor{bestgreen}4.64 & 2.50 & 3.00 & 3.29 &
\cellcolor{bestgreen}3.39 & 2.29 & 2.77 & 2.63 & max \\
\hline
\textbf{Norm.\ SP length} &
\cellcolor{bestgreen}1.05 & 1.06 & 1.09 & 1.12 &
\cellcolor{bestgreen}1.03 & 1.15 & 1.12 & 1.14 &
\cellcolor{bestgreen}1.04 & 1.12 & 1.06 & 1.09 & 1 \\
\hline
\end{tabular}
\end{table*}

\textbf{Env.~3.}
The proposed method leads across low-to-moderate fleet sizes under PIBT,
exceeding GSRM by $12.3\,\%$ at $|A|=50$ and by $3.5\,\%$ at $|A|=150$.
At $|A|=250$, GSRM marginally overtakes the proposed method by $1.0\,\%$;
at maximum fleet size ($|A|=300$), the proposed method recovers to
$1.2\,\%$ higher median throughput ($1.630$ vs.\ $1.611$~tasks/s). The
grid trails consistently ($9.1\,\%$ below at $|A|=300$), while random
sampling declines sharply above $|A|=200$, finishing $38.1\,\%$ below at
$|A|=300$. Under the space-time A$^*$ solver, the proposed method leads
all baselines at low fleet sizes but converges with the grid at
intermediate densities ($|A|=36$--$48$), before clearly separating at
higher fleet sizes: at $|A|=60$, the proposed method exceeds GSRM by
$13.7\,\%$, the grid by $7.8\,\%$, and random sampling by $25.5\,\%$.

\subsection{Graph-Theoretical Metrics}
\label{subsec:graph_metrics}
Node counts are comparable across all methods by design, as the minimum
inter-node distance constraint of Eq.~\eqref{eq:d-min-nodes} limits node
density uniformly. Edge counts are among the highest for the proposed method, reflecting denser connectivity from geometry-aware
node placement. No ideal value is defined for node and
edge counts, as a higher node and edge count increases redundancy while also
expanding the search space.

The proposed method achieves the highest mean node and edge connectivity
between all station pairs with $T_{ij} > 0$ in all environments
(Tab.~\ref{tab:graph_metrics}). The advantage is most pronounced in Env.~2,
where mean node and edge connectivity reach $3.86$ and $4.64$, compared to
$2.39$/$2.50$ for GSRM, $2.89$/$3.00$ for the grid, and $2.82$/$3.29$ for
random sampling. In Env.~3, GSRM exhibits the lowest connectivity despite
comparable throughput, suggesting that in large open environments
inter-station connectivity is less decisive than in constrained layouts
such as Env.~2, where the advantage translates into a throughput gain of
more than $20\,\%$ over GSRM.

The proposed method achieves the lowest normalized mean shortest path
length in all three environments ($1.05$, $1.03$, $1.04$), approaching
the geometric optimum of~$1$ as a direct consequence of placing nodes at
convex corner points of $C_{\text{free}}$. The grid and random sampling
incur penalties of up to $1.09$ and $1.14$; GSRM reaches $1.15$ in
Env.~2. The combination of high connectivity and near-optimal path lengths
provides the structural basis for the consistent throughput advantages
observed across both solvers and all three environments.

\subsection{Pruning Analysis}
\label{subsec:pruning_analysis}
Comparing the pruned and full roadmaps of the proposed method across all
three environments under PIBT, pruning reduces nodes and edges by $21\,\%$
in Env.~1, $12$--$13\,\%$ in Env.~2, and $55$--$58\,\%$ in Env.~3,
while redundancy is largely preserved. Throughput
differences remain below $1.5\,\%$ at low-to-moderate fleet sizes; in
Env.~3 the pruned roadmap marginally outperforms the full roadmap at low
densities ($+1.2\,\%$ at $|A|=50$), as the second Delaunay triangulation
run can introduce improved connectivity. At maximum fleet size,
throughput reductions of $6.5\,\%$ (Env.~1) and $9.0\,\%$ (Env.~3)
reflect the tradeoff between structural compactness and routing
flexibility at high agent densities; in Env.~2 the pruned roadmap
retains a slight advantage. Results for the space-time A$^*$ solver and
the baseline methods show comparable behavior.

\section{Conclusion}
\label{sec:conclusion}

This paper presents a continuous-space roadmap generation method for
mobile robot fleet routing that places nodes at station interaction
points and convex corner points of the free space, expands coverage
via local grid expansion, enforces minimum inter-node and node-edge
distance constraints derived from robot dimensions, and applies
transport demand-driven $K$-shortest path pruning to reduce graph
complexity while preserving routing redundancy.

Evaluation across three intralogistics environments using two
independent MAPD solvers demonstrates consistent throughput improvement
over all baselines. Under PIBT, the proposed method exceeds GSRM by
$5.4\,\%$ in Env.~1, leads all baselines by at least $10.4\,\%$ in
Env.~2, and achieves $1.2\,\%$ higher throughput than GSRM in Env.~3,
each at maximum fleet size. The space-time A$^*$ solver confirms these
results, with margins of at least $1.0\,\%$, $21.7\,\%$, and $7.8\,\%$
in Environments~1,~2, and~3, respectively. Graph-theoretical analysis
confirms the highest inter-station connectivity and near-optimal
normalized path lengths of $1.03$--$1.05$ across all environments at
comparable roadmap complexity.

Future work will introduce a traffic zone concept to mitigate congestion
at high agent density, benchmark the pruning strategy against
alternative roadmap reduction approaches across multiple MAPD solvers,
and validate the method in physical robot deployments.

\bibliography{references}
\bibliographystyle{IEEEtran}

\end{document}